\documentclass[letterpaper]{article} % DO NOT CHANGE THIS
\usepackage{aaai2027} % switched from submission to final to show author info
\usepackage[hyphens]{url} % DO NOT CHANGE THIS
\usepackage{graphicx} % DO NOT CHANGE THIS
\usepackage{natbib} % DO NOT CHANGE THIS AND DO NOT ADD OPTIONS
\usepackage{caption} % DO NOT CHANGE THIS AND DO NOT ADD OPTIONS
\usepackage{algorithm}
\usepackage{algorithmic}
\usepackage{amsmath}
\usepackage{booktabs}
\usepackage{svg}

\newcommand{\method}{FractureFields}

\title{\method: Contact-Aware Binary Multi-Field Transfer for Fractured 3D Gaussian Simulation}

\author{%
  Jianchen Wang,
  Runyang Qu,
  Fei Li
}
\affiliations{%
  School of Computer Science and Technology, Xidian University
}

\begin{document}
\maketitle

\begin{abstract}
Physics-integrated 3D Gaussian representations enable the simulation of image-reconstructed assets directly as particles; however, current Gaussian–MPM pipelines maintain a single Eulerian velocity field even after fracture. When disconnected fragments share interpolation support, they still write to and read from the same grid nodes, resulting in cross-fragment momentum leakage that manifests as residual adhesion and non-physical stretching. We present \method, a topology-adaptive transfer for fractured 3D Gaussian objects. After a structural event assigns persistent fragment identities, \method constructs fragment-specific mass and momentum fields in a single P2G pass, advances each field independently, and performs a field-aware G2P update so that particles only sample their own fragment’s grid state. To manage re-contact, we add a momentum-conserving contact projection that applies equal and opposite normal impulses only when two fragment fields are approaching, thus preserving free separation otherwise. Experiments on reconstructed scenes and a controlled re-contact benchmark demonstrate that fragment-conditioned routing eliminates realized cross-fragment mixing by construction, while contact projection reduces interpenetration during collisions without reintroducing residual coupling. Overall, we argue that post-fracture simulation should treat structural disconnection as a change in local dynamical state rather than merely a change in constitutive stress.
\end{abstract}

% ================================================================
% FIGURE 1 RESPONSIBILITY:
% Show the real failure, the proposed behavior, and one node-level inset.
% Replace this box with a cropped PDF/PNG exported outside LaTeX.
% ================================================================
\begin{figure}[t]
    \centering
    \includegraphics[width=\columnwidth]{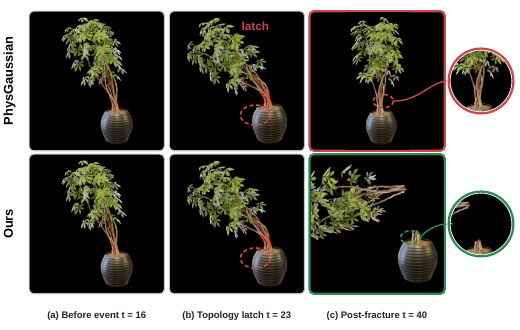}

    \caption{\textbf{Qualitative comparison on the Ficus scene.}
    PhysGaussian and our method are evaluated at matched time steps.
    After fracture, the shared-field baseline exhibits residual fragment
    adhesion, whereas our fragment-conditioned transfer enables persistent
    separation. The circular insets magnify the fracture region.}
    \label{fig:teaser}
\end{figure}

\section{Introduction}

Photorealistic 3D reconstruction is increasingly expected to support not only novel-view rendering, but also physically plausible response to user intervention. 3D Gaussian Splatting (3DGS) offers an explicit scene representation that is efficient to render \cite{kerbl2023gaussian}. Recent physics-integrated variants go further by treating Gaussians (and their interior samples) as simulation particles, enabling reconstructed assets to deform and move under forces without introducing a separate proxy mesh \cite{xie2024physgaussian,cai2024gic,jiang2024vrgs,feng2024gaussiansplashing}. This unified view is appealing for interactive content creation, embodied simulation, and controllable scene editing.

However, most Gaussian--MPM pipelines retain a topology-continuous \emph{dynamical state}. In MLS--MPM, particles transfer mass and momentum to a shared Eulerian grid where each node stores a single velocity. This representation is consistent for connected continua, but it becomes inconsistent after fracture: even if an event detector declares the body disconnected, particles from different fragments may still overlap in interpolation support, write to the same grid nodes, and then read back the same mass-averaged nodal velocity. In other words, the particle topology changes, but the grid state remains single-valued.

This mismatch induces \emph{cross-fragment momentum leakage} through P2G/G2P normalization. At shared-support nodes, nodal averaging erases the intended velocity discontinuity, so fragments that should separate can appear to stick, move synchronously, or stretch across an already-open crack. Importantly, stress softening or interface-band stress release can reduce traction transmission, but it cannot remove this kinematic coupling as long as both fragments are forced to sample the same nodal velocity. A brute-force fix is to run a full P2G/Grid/G2P cycle independently per fragment, but that repeats global work and complicates re-contact.

Our key idea is to make the Eulerian state multi-valued \emph{only where needed}. We introduce \method, a topology-adaptive binary multi-field transfer for fractured 3D Gaussian simulation. After a structural event assigns persistent fragment identities to particles, \method routes each particle to a fragment-specific nodal mass and momentum field within a single P2G traversal. Nodes influenced by a single fragment reduce to the standard MLS--MPM update; only shared-support nodes maintain two independent velocities. Identity-preserving G2P then ensures each particle samples only its own fragment field, eliminating post-fracture nodal averaging by construction.

A discontinuous velocity state is not sufficient on its own because fragments may later re-contact. \method therefore adds a lightweight, momentum-conserving contact projection that activates only when the two fragment fields are approaching along the latched interface normal: separating fields receive no impulse, while colliding fields receive equal and opposite normal impulses. The transfer is agnostic to how fracture is detected; we use oracle topology in controlled studies and a stress-guided event provider for reconstructed scenes.

We validate \method with experiments that isolate (i) transfer-induced mixing under low and high shared-support overlap, (ii) runtime and memory trade-offs relative to shared transfer and global per-fragment passes, and (iii) re-contact behavior in a controlled fragment--fragment benchmark. In particular, Table~\ref{tab:transfer_validation} reports counterfactual versus realized mixing (showing structural zero realized mixing under identity-preserving routing), Table~\ref{tab:transfer_efficiency_v3} reports timing and memory overheads, and Table~\ref{tab:scene_interaction} evaluates re-contact penetration and momentum consistency.

Our contributions are:
\begin{itemize}
    \item We identify post-fracture coupling in Gaussian--MPM as cross-fragment momentum leakage caused by a mismatch between disconnected particle topology and a single-valued grid velocity field.
    \item We propose a topology-adaptive binary multi-field transfer with identity-preserving routing (P2G/G2P) that constructs fragment-specific nodal states in one pass and yields structural zero realized mixing.
    \item We develop an approach-only, momentum-conserving contact projection between fragment fields and validate re-contact behavior without reintroducing cross-fragment averaging.
\end{itemize}

\section{Related Work}

We situate \method at the intersection of physics-aware 3D Gaussian representations and multi-field MPM formulations for fracture and contact.

\paragraph{Dynamic neural and Gaussian scene representations.}
Neural radiance fields enable photorealistic novel-view synthesis with continuous volumetric representations~\cite{mildenhall2020nerf}, and later work extends them to time-dependent scenes~\cite{pumarola2021dnerf}. 3D Gaussian Splatting provides an explicit, efficiently rendered set of anisotropic primitives~\cite{kerbl2023gaussian}. Dynamic Gaussian variants preserve primitive identity for tracking~\cite{luiten2024dynamic3d} and support time-varying deformation for real-time dynamic rendering~\cite{wu2024fourDGS}. These methods primarily reconstruct or replay observed motion; our focus is on physically simulating \emph{unseen} interventions, especially after topology changes.

\paragraph{Physics-aware scene representations.}
A growing line of work couples learned scene representations with physical simulation. PAC-NeRF couples NeRFs with continuum dynamics for system identification~\cite{li2023pacnerf}. PhysGaussian uses Gaussians as both rendering primitives and MPM particles~\cite{xie2024physgaussian}, while related systems infer material properties~\cite{cai2024gic}, embed differentiable particle simulation into reconstruction~\cite{ni2024phyrecon}, or distill generative priors into material fields and MPM dynamics~\cite{zhang2024physdreamer}. Other approaches target interactive deformation and solid--fluid coupling~\cite{jiang2024vrgs,feng2024gaussiansplashing}, decouple contacted surfaces for scene interaction~\cite{wang2025decoupledgaussian}, or integrate interior synthesis with damage and fracture models~\cite{huang2026gaussianfluent}. Recent work explores implicit MPM integration and accelerated interior completion or optimization~\cite{cao2026iphysgaussian,ma2026fastphysgs}, hybrid real--synthetic simulation~\cite{alfonso2026physplat}, image- or video-conditioned dynamics~\cite{tan2026physmotion,jiang2026physho}, and occlusion-robust multi-object decoupling~\cite{dong2026occlusion}. In contrast to these acquisition- and modeling-focused advances, we study a specific simulation failure mode that arises \emph{after} a structural event: particle--grid transfer can re-couple fragments through nodal averaging unless the grid state itself becomes fragment-conditioned.

\paragraph{Multi-field MPM for fracture, transfer, and contact.}
The material point method (MPM) evolves Lagrangian particles through a temporary Eulerian grid~\cite{sulsky1994particle} and has become a standard tool for large-deformation elastoplastic simulation in graphics~\cite{stomakhin2013snow,jiang2016mpmcourse}. APIC and MLS--MPM refine particle--grid transfer by enriching particle velocity with locally affine motion and providing a compact weak-form view with two-way rigid coupling~\cite{jiang2015apic,hu2018mlsmpm}. When topology changes, however, a single-valued nodal state is no longer sufficient: explicit-crack and multimaterial MPM introduce multiple velocity fields near discontinuities and reconcile their nodal momenta across cracks or interfaces~\cite{nairn2003cracks,guo2006fracture,nairn2013interfaces}. Complementary lines of work construct contact pairs from field gradients~\cite{homel2017fieldgradient} or represent fracture and fragmentation through phase fields, continuum damage, and cohesive zones, together with evolving multi-body contact handling~\cite{kakouris2019phasefield,wolper2019cdmpm,zeng2023phasefield,wolper2020anisompm,xiao2021dpmpm,crook2026cohesive}. Contact algorithms in MPM likewise use body-conditioned nodal quantities to distinguish collision from free separation and to preserve conservation during impulses~\cite{bardenhagen2001contact,pan2008multimesh,tupek2021contact,menager2026contact}. We build on these established mechanics principles but target a different setting: reconstructed Gaussian objects simulated with Gaussian--MPM. Our contribution is a topology-adaptive binary realization that makes the grid state fragment-conditioned only where supports overlap, preserves identity through field-aware G2P to prevent cross-fragment fallback, and applies a normal-only impulse only to approaching fields.

\section{Problem Formulation and Failure Analysis}

This section formalizes why a standard single-velocity Eulerian grid becomes inconsistent after fracture in Gaussian--MPM, and isolates the resulting transfer-induced coupling between disconnected fragments.

\subsection{Gaussian--MPM Transfer}

At substep $n$ (time $t^n$), particle $p$ carries position $\mathbf{x}_p^n$, velocity $\mathbf{v}_p^n$, mass $m_p$, volume $V_p$, affine velocity $\mathbf{C}_p^n$, deformation gradient $\mathbf{F}_p^n$, and stress $\boldsymbol{\sigma}_p^n$. Following MLS--MPM~\cite{hu2018mlsmpm}, the standard particle-to-grid (P2G) transfer aggregates contributions from all particles into a single nodal mass and momentum:
\begin{equation}
m_i^n=\sum_p w_{ip}m_p,
\label{eq:single_mass_v2}
\end{equation}
\begin{equation}
\mathbf{p}_i^n=
\sum_p w_{ip}m_p\left(\mathbf{v}_p^n+\mathbf{C}_p^n(\mathbf{x}_i-\mathbf{x}_p^n)\right)
-\Delta t\sum_p V_p\boldsymbol{\sigma}_p^n\nabla w_{ip}.
\label{eq:single_momentum_v2}
\end{equation}
After applying gravity and boundary conditions, each node stores a single velocity
\begin{equation}
\mathbf{v}_i^{n+1}=\mathcal{B}\!\left(
\frac{\mathbf{p}_i^n}{m_i^n}+\Delta t\mathbf{g}\right),
\label{eq:single_velocity_v2}
\end{equation}
which is interpolated back to all particles in the node's interpolation support. This single-field grid state is consistent as long as the simulated body remains topologically connected.

\subsection{Single-Field Topology Mismatch}

A key observation is that \emph{persistent fragment identities alone do not imply a discontinuous grid state}. If the grid continues to store a single velocity per node, then any node that receives contributions from multiple fragments will necessarily produce a mass-averaged velocity that couples their motion.

Suppose a structural event at time $t_{\mathrm{latch}}$ assigns each particle a persistent fragment identity $o_p\in\{0,1\}$. Let $(m_i^k,\mathbf{p}_i^k)$ denote the nodal mass and momentum accumulated from fragment $k$. If the simulator continues to use a shared grid with one velocity per node, the nodal velocity is computed from the total mass and momentum,
\begin{equation}
\bar{\mathbf{v}}_i=
\frac{\mathbf{p}_i^0+\mathbf{p}_i^1}
     {m_i^0+m_i^1}.
\label{eq:shared_velocity}
\end{equation}
We call node $i$ a \emph{shared-support} node if both $m_i^0$ and $m_i^1$ are non-negligible. At such nodes, particles on opposite sides of the latched interface read back the same mass-averaged velocity, creating an unphysical kinematic coupling after fracture. This situation is common in Gaussian--MPM because reconstruction yields dense particle sampling, while interpolation stencils remain fixed across the latch event.

Notably, setting the interface-band stress to zero only modifies the internal-force term in Eq.~\ref{eq:single_momentum_v2}; it does not change the shared normalization in Eq.~\ref{eq:shared_velocity}, and therefore cannot restore a velocity discontinuity.

To quantify the coupling induced by shared normalization, we measure the discrepancy:
\begin{equation}
E_{\mathrm{mix}}=
\frac{
\sum_{i,k}m_i^k
\left\|\bar{\mathbf{v}}_i-\mathbf{v}_i^k\right\|_2^2
}{
\sum_{i,k}m_i^k+\epsilon
},
\qquad
\mathbf{v}_i^k=\frac{\mathbf{p}_i^k}{m_i^k}.
\label{eq:mixing}
\end{equation}
The numerator can also be interpreted as a reduced-mass weighting of $\|\mathbf{v}_i^0-\mathbf{v}_i^1\|_2^2$ over shared-support nodes. The measure is zero when every occupied node is influenced by only one fragment, and becomes positive exactly when disconnected fragments require different local velocities at overlapping stencils.

\subsection{Desired Post-Fracture State}

A physically meaningful post-fracture update should satisfy three consistency requirements. (i) \emph{Impulse-free latching:} the latch event changes only the routing/labeling, and particle position, velocity, affine state, and deformation are inherited continuously. (ii) \emph{Fragment-conditioned accumulation:} particle mass remains attached to its persistent identity and is accumulated into fragment-specific nodal masses. (iii) \emph{Identity-preserving evolution and return transfer:} the same identity must condition nodal momentum, velocity, APIC affine motion, and deformation updates, and particles must read back from their own fragment state. External forces and subsequent contact may change momentum, but the update should not reintroduce cross-fragment velocity averaging.

In the next section, we show that these requirements can be met by (a) routing particles into fragment-conditioned nodal fields during P2G, (b) evolving those fields independently except for contact, and (c) enforcing identity-preserving G2P so particles never fall back to a mixed state.

% ================================================================
% FIGURE 2 RESPONSIBILITY:
% Left: failure derivation. Right: method data flow.
% The same particle positions and grid must be used on both sides.
% ================================================================
\begin{figure*}[t]
    \centering
    \includegraphics[width=0.98\textwidth]{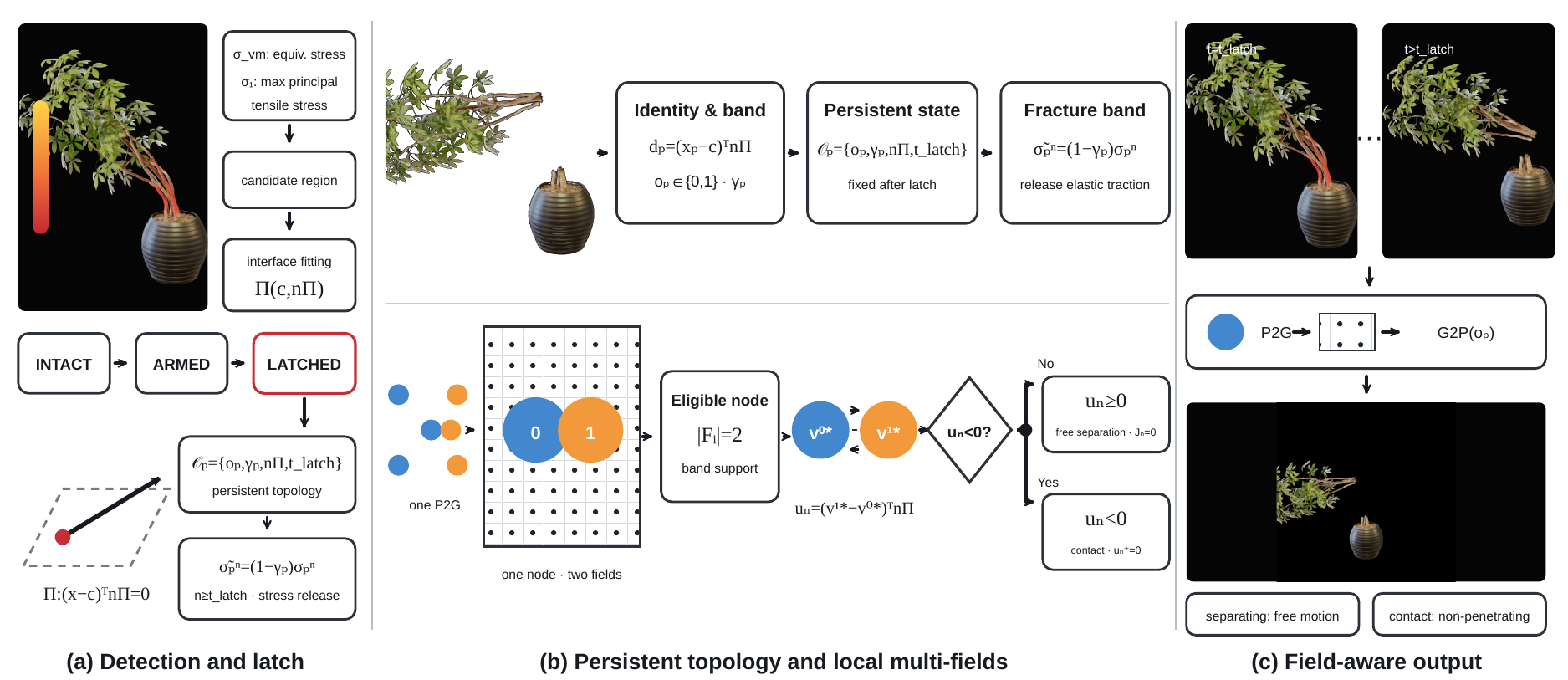}
    \caption{\textbf{Overview of the proposed fracture-aware simulation pipeline.}
    (a) A stress-guided event detector localizes the fracture interface and
    temporally latches persistent topology. 
    (b) Fragment identities route particles into separate local mass and
    momentum fields during a single P2G pass.
    (c) The fields evolve independently during separation and activate contact
    handling only when approaching, after which field-aware G2P returns each
    particle from its corresponding field.}
    \label{fig:method}
\end{figure*}

\section{\method}

\subsection{Overview}

\method changes only the post-fracture particle--grid transfer and contact handling; it leaves the constitutive model, Gaussian updates, and rendering pipeline untouched. We assume a persistent binary topology descriptor
\begin{equation}
\mathcal{O}=\{o_p,\gamma_p,\mathbf{n}_{\Pi},t_{\mathrm{latch}}\},
\end{equation}
where $o_p\in\{0,1\}$ is the fragment identity, $\gamma_p\in\{0,1\}$ marks particles in a narrow interface band, $\mathbf{n}_{\Pi}$ is the latched interface normal, and $t_{\mathrm{latch}}$ is the latch time.

For each simulation substep with $t^n\ge t_{\mathrm{latch}}$, we (i) perform fragment-conditioned P2G to build fragment-specific nodal fields, (ii) independently normalize and advance each active field on the grid (gravity and standard grid-side operators, with internal forces accounted for in P2G), (iii) optionally project approaching fields for contact at shared-support nodes near the interface, and (iv) perform identity-preserving G2P so each particle samples only its own fragment field.

The transfer is agnostic to how $\mathcal{O}$ is obtained: controlled experiments specify it analytically, while reconstructed-scene experiments use a stress-guided provider that estimates and then temporally latches an interface before assigning persistent identities to both visible and interior particles. Detection details are deferred to the supplementary material because the transfer formulation does not depend on a particular fracture criterion.

\subsection{Latching and Interface Treatment}

The latch event changes only the transfer topology and should not introduce an artificial impulse. Denoting particle states immediately before and after latching by $-$ and $+$, we inherit kinematics and deformation,
\begin{equation}
\mathbf{x}_p^+=\mathbf{x}_p^-,\quad
\mathbf{v}_p^+=\mathbf{v}_p^-,\quad
\mathbf{C}_p^+=\mathbf{C}_p^-,\quad
\mathbf{F}_p^+=\mathbf{F}_p^-.
\label{eq:state_inheritance_v2}
\end{equation}
with $m_p^+=m_p^-$ and $V_p^+=V_p^-$. The only new persistent state is the fragment label $o_p$, assigned from the signed distance to the latched interface. This choice preserves total momentum at the event,
\begin{equation}
\begin{aligned}
\mathbf{P}^{+} &= \sum_p m_p\mathbf{v}_p^{+}=\sum_p m_p\mathbf{v}_p^{-}=\mathbf{P}^{-},\\
\mathbf{P}_k^{+} &= \sum_{p:o_p=k}m_p\mathbf{v}_p^{-}.
\end{aligned}
\label{eq:momentum_inheritance_v2}
\end{equation}
so the pre-event momentum is partitioned into persistent fragments without modification.

To suppress spurious traction transfer across the frozen interface, we optionally release stress for interface-band particles. For $t^n\ge t_{\mathrm{latch}}$, the stress used by P2G is
\begin{equation}
\widetilde{\boldsymbol{\sigma}}_p^n=(1-\gamma_p)\boldsymbol{\sigma}_p^n.
\label{eq:stress_release_v2}
\end{equation}
Particles outside the band retain their constitutive response, while band particles contribute no elastic traction across the interface. This stress release targets force transmission; \method separately targets the post-fracture kinematic coupling caused by shared velocity normalization.

\subsection{Algorithm Summary}

Algorithm~\ref{alg:fracturefields} summarizes the post-latch update executed at each simulation substep.

\begin{algorithm}[t]
\caption{\method post-latch substep update}
\label{alg:fracturefields}
\begin{algorithmic}
\REQUIRE Particles with states $(\mathbf{x}_p,\mathbf{v}_p,\mathbf{C}_p,\mathbf{F}_p,m_p,V_p,\boldsymbol{\sigma}_p)$ and persistent topology labels $(o_p,\gamma_p,\mathbf{n}_{\Pi},t_{\mathrm{latch}})$.
\STATE (Optional) Apply interface-band stress release (Eq.~\ref{eq:stress_release_v2}).
\STATE Reset fragment-conditioned grid accumulators for both fields $k\in\{0,1\}$.
\FOR{each particle $p$}
    \STATE Accumulate fragment-conditioned P2G to field $k\!=\!o_p$ (Eq.~\ref{eq:multifield_p2g}).
    \STATE Accumulate interface-band mass $m_{i,\mathrm{band}}^k$ (Eq.~\ref{eq:band_mass}) for contact gating.
\ENDFOR
\FOR{each node $i$ and each active field $k$}
    \STATE Normalize field velocity and apply gravity.
\ENDFOR
\FOR{each shared-support node $i$}
    \STATE If both fields are band-supported ($m_{i,\mathrm{band}}^k>\epsilon_{\mathrm{band}}$) and approaching along $\mathbf{n}_{\Pi}$, apply contact projection (Eqs.~\ref{eq:impulse}--\ref{eq:contact_update}).
\ENDFOR
\FOR{each node $i$ and each active field $k$}
    \STATE Apply boundary operators to the (optionally projected) field velocity.
\ENDFOR
\FOR{each particle $p$}
    \STATE Identity-preserving G2P from field $k\!=\!o_p$ (Eq.~\ref{eq:field_g2p}).
    \STATE Update particle state using the same field velocities.
\ENDFOR
\end{algorithmic}
\end{algorithm}

\subsection{Topology-Adaptive Multi-Field Transfer}

For each node $i$, we store up to two fragment-conditioned fields and define the active field set
\begin{equation}
\mathcal{F}_i=\{k\in\{0,1\}\mid m_i^k>\epsilon_m\}.
\end{equation}
A node is \emph{shared-support} if $|\mathcal{F}_i|=2$ (equivalently, $m_i^0>\epsilon_m$ and $m_i^1>\epsilon_m$).

\textbf{Fragment-conditioned P2G.} Each particle writes \emph{only} to the field indexed by its persistent identity $o_p$:
\begin{equation}
m_i^k=\sum_{p:o_p=k}w_{ip}m_p,
\end{equation}
\begin{equation}
\begin{aligned}
\mathbf{p}_i^k &= \sum_{p:o_p=k} w_{ip}m_p\left(\mathbf{v}_p+\mathbf{C}_p(\mathbf{x}_i-\mathbf{x}_p)\right)\\
&\quad-\Delta t\sum_{p:o_p=k} V_p\widetilde{\boldsymbol{\sigma}}_p\nabla w_{ip}.
\end{aligned}
\label{eq:multifield_p2g}
\end{equation}
We evaluate Eq.~\ref{eq:multifield_p2g} in a single particle traversal. In contrast, global per-fragment passes clear and process the full grid once per fragment.

\textbf{Independent field update.} Each active field is normalized independently,
\begin{equation}
\mathbf{v}_i^{k,*}=
\frac{\mathbf{p}_i^k}{m_i^k}
+\Delta t\,\mathbf{g},
\end{equation}
and we then apply the same grid-side operators used by the single-field solver (e.g., boundary conditions and internal-force updates) independently to each active field. If $|\mathcal{F}_i|=1$, the node reduces to the standard MLS--MPM update.

\textbf{Routing invariant.} Fragment-conditioned P2G and identity-preserving G2P enforce an invariant: for all substeps with $t^n\ge t_{\mathrm{latch}}$, particle $p$ can influence and sample only the grid field indexed by $o_p$. If a node does not activate field $k$ (i.e., $m_i^k\le\epsilon_m$), its contribution is treated as zero and we do not fall back to the other field, avoiding re-coupling by interpolation.

\textbf{Implication (structural zero mixing).} Under this invariant, the realized transfer update cannot compute a shared, mass-averaged nodal velocity across fragments at shared-support nodes. Consequently, the realized mixing metric in Eq.~\ref{eq:mixing} is identically zero for the rollout that uses \method, even if a counterfactual shared-grid average applied to the same particle states would yield $E_{\mathrm{mix}}>0$.

The transfer is mass consistent for each fragment. We measure the residual
\begin{equation}
E_m^k=
\frac{\left|\sum_i m_i^k-\sum_{p:o_p=k}m_p\right|}
{\sum_{p:o_p=k}m_p+\epsilon}.
\label{eq:mass_error_v2}
\end{equation}
Under a complete interpolation stencil, partition of unity gives $E_m^k=0$ up to floating-point error. Together, Eqs.~\ref{eq:state_inheritance_v2} and \ref{eq:multifield_p2g} define the transition from one connected velocity field before fracture to two fragment-conditioned fields after fracture.

\subsection{Contact-Aware Grid Projection}

Independent fields remove erroneous adhesion but also remove the implicit contact response of a shared grid. We therefore introduce an explicit projection, applied only at shared-support nodes near the latched interface.

\textbf{Interface gating.} We compute an interface-band mass per node and field,
\begin{equation}
m_{i,\mathrm{band}}^k=\sum_{p:o_p=k}\gamma_p\,w_{ip}m_p,
\label{eq:band_mass}
\end{equation}
and call field $k$ \emph{band-supported} at node $i$ if $m_{i,\mathrm{band}}^k>\epsilon_{\mathrm{band}}$. We use this gate to avoid treating incidental overlap away from the fracture surface as self-contact, and to preserve free separation except in the vicinity of the latched interface. In all experiments, we choose $\epsilon_m$ and $\epsilon_{\mathrm{band}}$ as small fractions of typical occupied-node masses and set $\epsilon_{\mathrm{band}}$ on the same order as $\epsilon_m$. The small constant $\epsilon$ in Eq.~\ref{eq:mixing} is used solely to avoid division by zero in edge cases involving empty-nodes. Exact numeric values are provided in the supplementary material.

\textbf{Normal-only projection.} At a shared-support node with both fields band-supported, we compute the normal relative speed
\begin{equation}
u_n=(\mathbf{v}_i^{1,*}-\mathbf{v}_i^{0,*})^\top\mathbf{n}_{\Pi}.
\end{equation}
If $u_n\ge 0$, the fields are separating and we apply no impulse. If $u_n<0$, we apply a frictionless normal impulse
\begin{equation}
J_n=-\frac{(1+e)u_n}{1/m_i^0+1/m_i^1},
\qquad e=0,
\label{eq:impulse}
\end{equation}
which yields the projected velocities
\begin{equation}
\mathbf{v}_i^{0,+}=\mathbf{v}_i^{0,*}-\frac{J_n}{m_i^0}\mathbf{n}_{\Pi},
\qquad
\mathbf{v}_i^{1,+}=\mathbf{v}_i^{1,*}+\frac{J_n}{m_i^1}\mathbf{n}_{\Pi}.
\label{eq:contact_update}
\end{equation}
The equal and opposite impulses preserve node-level linear momentum,
\begin{equation}
m_i^0\mathbf{v}_i^{0,+}+m_i^1\mathbf{v}_i^{1,+}=m_i^0\mathbf{v}_i^{0,*}+m_i^1\mathbf{v}_i^{1,*}.
\end{equation}
Because the projection is normal-only, tangential relative motion is unchanged in the frictionless setting. In our implementation, we apply the original boundary operators to each active field after contact projection, so boundary handling is consistent with the projected velocities.

\begin{figure*}[t]
    \centering
    \includegraphics[width=0.98\textwidth]{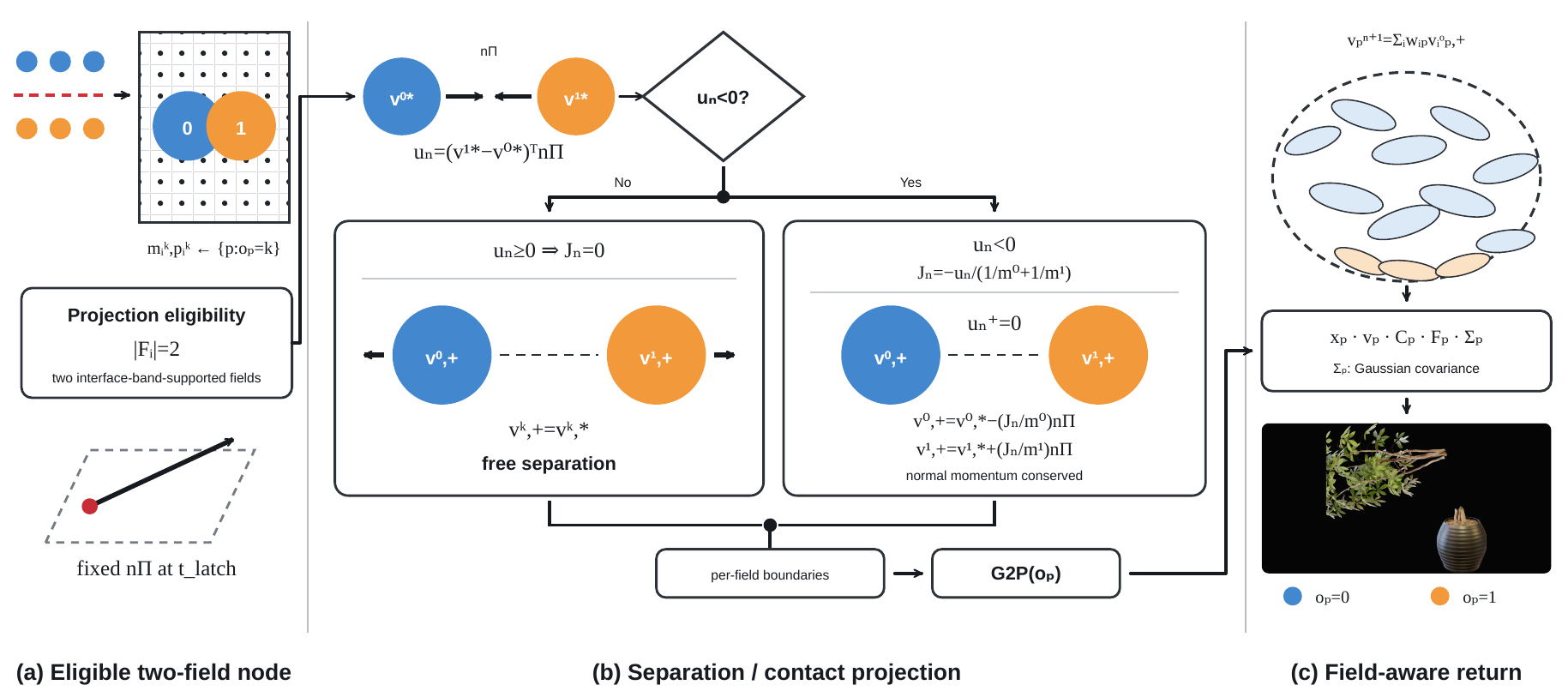}
    \caption{\textbf{Contact-aware local multi-field update.}
    (a) Persistent particle identities divide the fractured object into two
    fields across the latched interface.
    (b) Fragment-aware P2G constructs separate nodal velocities. Separating
    fields receive no impulse, while approaching fields receive equal and
    opposite normal impulses that preserve node-level linear momentum.
    (c) Field-aware G2P reads the field associated with each particle identity,
    preserving the post-fracture velocity discontinuity.}
    \label{fig:contact_algorithm}
\end{figure*}

\subsection{Field-Aware G2P and Properties}

Particle $p$ reads only the field indexed by its persistent identity,
\begin{equation}
\mathbf{v}_p^{n+1}=\sum_i w_{ip}\mathbf{v}_i^{o_p,+}.
\label{eq:field_g2p}
\end{equation}
and the same field velocities are used to update $\mathbf{C}_p$, the velocity gradient, $\mathbf{F}_p$, Gaussian covariance, and particle position. We do not fall back to the opposite field, since any cross-field fallback would reintroduce the coupling that multi-field P2G is designed to remove.

The affine and deformation updates use the same fragment index:
\begin{equation}
\mathbf{C}_p^{n+1}=\frac{4}{\Delta x^2}\sum_i w_{ip}\mathbf{v}_i^{o_p,+}(\mathbf{x}_i-\mathbf{x}_p^n)^{\top},
\label{eq:field_apic_v2}
\end{equation}
\begin{equation}
\mathbf{F}_p^{n+1}=\left(\mathbf{I}+\Delta t\sum_i \mathbf{v}_i^{o_p,+}(\nabla w_{ip})^{\top}\right)\mathbf{F}_p^n.
\label{eq:field_deformation_v2}
\end{equation}
Consequently, fragment separation influences not only translational motion but also the affine and deformation states that drive Gaussian covariance.

\paragraph{Properties and scope.}
\method has three key properties: it is single-field consistent (assigning all particles to one identity recovers standard MLS--MPM), it prevents post-fracture nodal averaging by enforcing identity-consistent routing across P2G and G2P, and its contact projection preserves node-level linear momentum via equal and opposite impulses.

Our implementation targets a persistent binary split with dense two-field storage. It executes a single particle traversal per substep (instead of a full grid pass per fragment) at the cost of storing two field states at shared-support nodes; extending the approach to sparse multi-fragment allocation is left for future work.

\section{Experiments}

\subsection{Experimental Setup and Evaluation Protocol}

We implement \method in the Warp-based PhysGaussian solver~\cite{xie2024physgaussian} and evaluate reconstructed Ficus and Bread assets. Ficus contains 171,553 MPM particles and is simulated for 44 frames with frame interval $0.04$~s; Bread contains 182,959 particles and is simulated for 90 frames with frame interval $0.01$~s. Structural events latch at frames 23 (Ficus) and 10 (Bread). The saved reference topology contains 164 interface pairs for Ficus and 906 for Bread. Unless stated otherwise, post-fracture statistics use frames 23--43 (Ficus) and 10--89 (Bread).

We compare four real-scene variants that isolate individual components. \emph{Baseline} is the original shared-field solver. \emph{Probe+Latch} adds the event detector and frozen fragment labels. \emph{Stress-Zero} additionally releases interface-band stress while keeping shared transfer. \emph{Full} enables fragment-conditioned transfer; contact projection is enabled only for interaction experiments. All variants use the same saved topology for diagnostics.

We evaluate three questions: (i) does fragment-conditioned transfer eliminate transfer-induced mixing at shared-support nodes, (ii) what are the runtime and memory trade-offs relative to shared transfer and per-fragment global passes, and (iii) does the method preserve both free separation and physically plausible re-contact.

\begin{figure*}[t]
    \centering
    \includegraphics[width=0.98\textwidth]{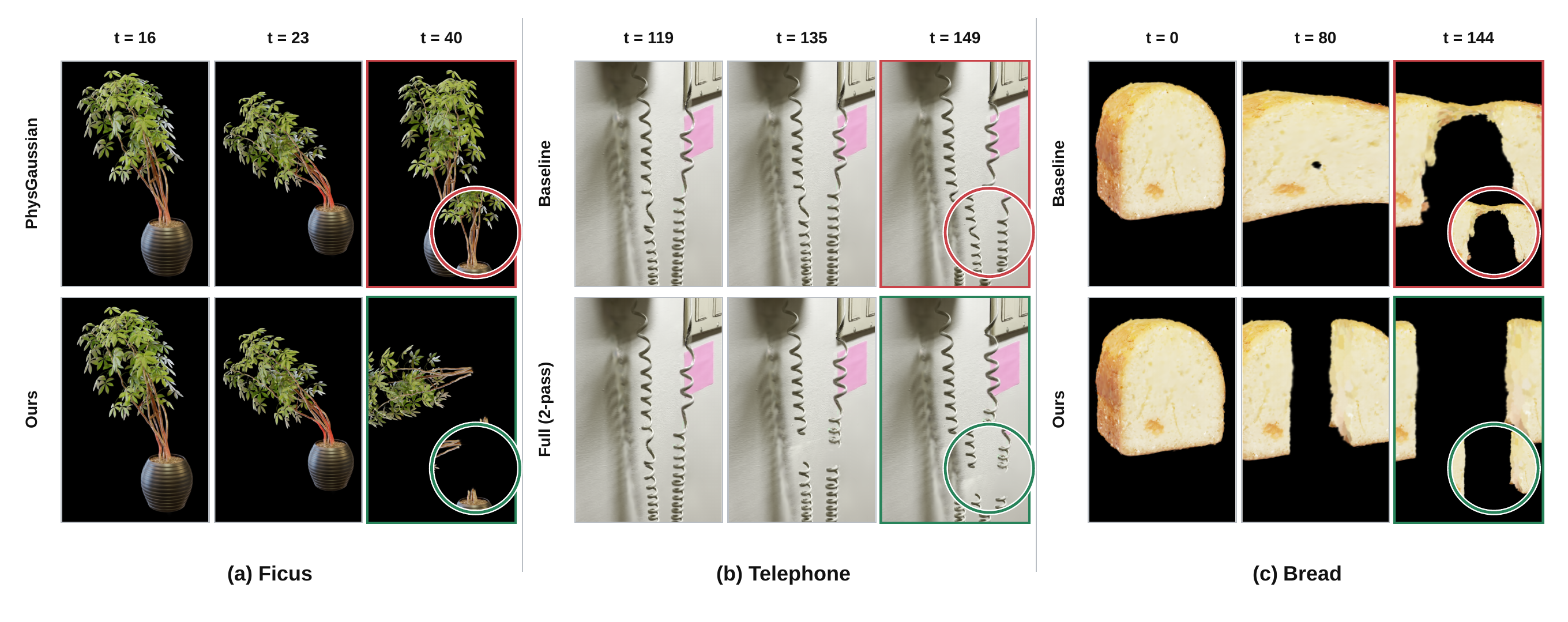}
    \caption{\textbf{Qualitative comparison across reconstructed datasets.}
    Rows correspond to different reconstructed scenes, while columns compare
    the shared-field baseline, Stress-Zero, and our full method at matched
    post-fracture time steps. Our fragment-conditioned transfer produces
    cleaner and persistent separation with less residual adhesion and
    fracture-band stringing.}
    \label{fig:multi_dataset_comparison}
\end{figure*}

Figure~\ref{fig:multi_dataset_comparison} compares the post-fracture
results across the reconstructed datasets at matched time steps.

\subsection{Validation of Fragment-Wise Transfer}

We first test whether fragment-conditioned transfer eliminates the velocity averaging that remains after stress release. Table~\ref{tab:transfer_validation} compares Stress-Zero and Full, which share the same event and stress-release components and differ only in their post-fracture transfer. Counterfactual mixing evaluates the shared-grid average on the current particle state; realized mixing records the transfer actually used by the rollout.

\begin{table*}[t]
\centering
\scriptsize
\begin{tabular}{llcccc}
\toprule
Scene & Transfer & $E_{\mathrm{mix}}^{\mathrm{cf}}$ mean/max & $E_{\mathrm{mix}}^{\mathrm{real}}$ mean/max & Shared (\%) & $\max E_m$ \\
\midrule
Ficus & Shared (Stress-Zero) & $5.156{\times}10^{-7}/7.637{\times}10^{-6}$ & $5.156{\times}10^{-7}/7.637{\times}10^{-6}$ & 2.280 & 0 \\
Ficus & Fragment-conditioned (Full) & $5.436{\times}10^{-5}/7.099{\times}10^{-4}$ & $0/0$ & 2.520 & 0 \\
Bread & Shared (Stress-Zero) & $1.637{\times}10^{-6}/6.721{\times}10^{-6}$ & $1.637{\times}10^{-6}/6.721{\times}10^{-6}$ & 96.580 & 0 \\
Bread & Fragment-conditioned (Full) & $3.424{\times}10^{-3}/5.346{\times}10^{-3}$ & $0/0$ & 96.420 & 0 \\
\bottomrule
\end{tabular}
\caption{\textbf{Post-fracture transfer diagnostics.} Mixing entries are post-latch mean/maximum over frames 23--43 for Ficus and 10--89 for Bread. For Full, realized mixing is a structural zero established by the P2G/G2P routing invariant. Mass residuals are zero at the exported precision.}
\label{tab:transfer_validation}
\end{table*}

\noindent\textbf{Summary.} Shared transfer realizes the counterfactual mixing error, while Full achieves zero realized mixing by construction even under very high shared-support ratios (Bread). This directly validates the routing invariant: once fragment identities condition both P2G and G2P, there is no path for cross-fragment velocity averaging to re-enter the update.

\subsection{Efficiency of Local Multi-Field Transfer}

We compare transfer cost against two baselines. \emph{Global Fragment Pass} performs a separate full-grid P2G/Grid/G2P update per fragment, eliminating mixing but repeating global work. \emph{Local Multi-Field} uses our fragment-conditioned transfer without contact projection, isolating the cost of storing and updating two local fields. \method additionally enables contact projection. Timing includes grid reset, P2G, grid normalization, optional contact projection, and G2P. Each branch uses warm-up iterations followed by repeated measurements in multiple trials.

\begin{table*}[t]
\centering
\small
\begin{tabular}{lcccc}
\toprule
Branch & Passes (P2G/G2P/frag) & Median (ms) & Relative & Grid-buffer footprint (MiB) \\
\midrule
Shared Field & 1/1/0 & 0.589 & 1.000 & 56.000 \\
Global Fragment Pass & 2/2/1 & 0.713 & 1.211 & 56.000 \\
Local Multi-Field & 1/1/0 & 0.707 & 1.202 & 112.000 \\
\method & 1/1/0 & 0.726 & 1.232 & 112.000 \\
\bottomrule
\end{tabular}
\caption{\textbf{Transfer efficiency.} Runtime is normalized to Shared Field and excludes rendering, probing, export, and diagnostic shadow fields. The memory column is the branch-specific grid-buffer footprint.}
\label{tab:transfer_efficiency_v3}
\end{table*}

\noindent\textbf{Summary.} Local multi-field transfer avoids a second global grid pass but requires storing two field states at shared-support nodes, doubling the branch-specific grid buffer. This cost profile matches the design goal of \method: pay extra memory only where stencils overlap, while keeping the per-substep particle traversal count unchanged.

\subsection{Component Ablation on Reconstructed Scenes}

Table~\ref{tab:real_ablation_v2} isolates the effect of event latching, interface-band stress release, and fragment-conditioned transfer. Opening is the final mean interface opening normalized by $\Delta x$; Reapp. is the maximum re-approach ratio; Stringing measures deformation anisotropy in the fracture band; and Sep. applies the same scene-specific separation rule to every method.

\begin{table*}[t]
\centering
\small
\begin{tabular}{llccccc}
\toprule
Scene & Method & Opening/$\Delta x$ & Reapp. & Stringing & $\max E_m$ & Sep. \\
\midrule
Ficus & Baseline & $-0.018$ & 0.000 & 0.060 & 0 & No \\
Ficus & Probe+Latch & $-0.018$ & 0.000 & 0.060 & 0 & No \\
Ficus & Stress-Zero & 0.083 & 0.000 & 0.208 & 0 & No \\
Ficus & Full & \textbf{7.788} & 0.018 & 0.069 & 0 & Yes \\
\midrule
Bread & Baseline & $-0.423$ & 0.053 & 0.276 & 0 & No \\
Bread & Probe+Latch & $-1.750$ & 0.086 & 0.258 & 0 & No \\
Bread & Stress-Zero & 4.026 & 0.116 & 1.191 & 0 & Yes \\
Bread & Full & \textbf{10.009} & 0.022 & 0.341 & 0 & Yes \\
\bottomrule
\end{tabular}
\caption{\textbf{Real-scene component ablation.} Opening is measured over frames 23--43 for Ficus and 10--89 for Bread, and Sep. uses the common $0.1\Delta x$ opening threshold. Only Full combines stable topology, fracture-band stress release, and fragment-conditioned transfer.}
\label{tab:real_ablation_v2}
\end{table*}

\noindent\textbf{Summary.} Probe+Latch does not separate, confirming that frozen labels alone do not resolve transfer coupling. Stress-Zero can enable separation but may introduce strong fracture-band stringing. Full separates reliably while reducing stringing and re-approach, consistent with the intended division of labor: stress release suppresses spurious traction, while fragment-conditioned transfer removes kinematic averaging at shared-support nodes.

\subsection{Post-Fracture Interaction and Re-Contact}

We test whether independently transferred fragments remain interactive after fracture. Panel~A in Table~\ref{tab:scene_interaction} evaluates fragment--scene interaction under gravity (Ficus falling onto a frictionless plane). Panel~B is a controlled unequal-mass fragment re-contact benchmark; the two branches start from the same state and differ only in whether contact projection is enabled.

\begin{table*}[t]
\centering
\small
\resizebox{\textwidth}{!}{%
\begin{tabular}{llccccc}
\toprule
\multicolumn{7}{l}{\textit{Panel A: Ficus static-plane interaction (gravity-enabled, fragment--plane contact)}} \\
\midrule
Method & $t_c^0/t_c^1$ & Pen./$\Delta x$ & max $|v_t|$ near plane & global residual$^\dagger$ & max post-contact $v_n$ & Task Success \\
\midrule
Stress-Zero & $0.181/0.470$ & $0.000$ & $0.799$ & $1.173$ & $3.495$ & Yes \\
Full & $0.276/0.297$ & $0.026$ & $0.325$ & $1.139$ & $3.015$ & No \\
\midrule
\multicolumn{7}{l}{\textit{Panel B: Controlled fragment re-contact (gravity-disabled, fragment--fragment contact)}} \\
\midrule
Method & first-contact $t$ & Pen./$\Delta x$ & max $|\Delta v_t|$ & CP mom.\ residual & $v_{\mathrm{rel},n}^{+}$ at contact & Task Success \\
\midrule
Multi-field, no contact & $0.095$ & $2.833$ & $0.000$ & $0.000$ & $0.600$ & No \\
\method & $0.095$ & $0.053$ & $0.000$ & $1.167{\times}10^{-7}$ & $1.192{\times}10^{-7}$ & Yes \\
\bottomrule
\end{tabular}}
\caption{\textbf{Post-fracture interaction (dual-panel).} Panel~A tests fragment--scene response under gravity. $^\dagger$The global residual is gravity-compensated but includes the external plane contact impulse; it is a scene-level diagnostic, not a pure internal momentum-conservation error. Panel~B tests fragment--fragment re-contact without gravity. Task Success requires $\max(\mathrm{penetration})/\Delta x<0.1$ and no identity crossing. Panel-specific diagnostics must not be compared across panels.}
\label{tab:scene_interaction}
\end{table*}

\noindent\textbf{Summary.} In controlled re-contact, contact projection sharply reduces penetration while preserving tangential motion and maintaining negligible momentum residual. This supports the separation-aware design: the multi-field transfer preserves discontinuous motion during separation, and the normal-only projection activates only under approach to reintroduce contact response without re-coupling fragment velocities. In the fragment--scene interaction test (Panel~A), Full does not satisfy Task Success under the strict identity-crossing criterion, indicating that robust fragment--scene contact remains an open issue beyond the controlled fragment--fragment setting.

\subsection{Material-Prior Initialization}

We report a supplementary check on initialization: a vision-language material prior provides simulator parameters that lead to numerically stable rollouts under the same geometry, loading, and fracture protocol as a manual initialization. In this table only, Opening/$\Delta x$ is the signed proxy $\Delta d_n/\Delta x$ measured over the available rollout window.

\begin{table*}[t]
\centering
\small
\begin{tabular}{llccccccc}
\toprule
Scene & Init. & $E$ & $\rho$ & $\sigma_R$ & Latch & Opening/$\Delta x$ & Sep. & Valid \\
\midrule
Ficus & VLM & $2.0{\times}10^{6}$ & 200 & $5.5{\times}10^{4}$ & 24 & $-2.045$ & No & Yes \\
Ficus & Manual & $2.0{\times}10^{6}$ & 200 & $5.0{\times}10^{4}$ & 23 & $-2.101$ & No & Yes \\
Bread & VLM & $2.5{\times}10^{3}$ & 200 & 55 & 9 & 78.432 & Yes & Yes \\
Bread & Manual & $2.0{\times}10^{3}$ & 200 & 50 & 10 & 78.180 & Yes & Yes \\
\bottomrule
\end{tabular}
\caption{\textbf{Material-prior initialization.} The VLM values are obtained from GaussianProperty outputs and anchor-guided simulator mapping; they are initial conditions rather than recovered real-world SI parameters. Opening/$\Delta x$ is a signed, run-level proxy used only for within-scene VLM/manual comparison.}
\label{tab:material_prior}
\end{table*}

\noindent\textbf{Summary.} The mapped prior produces stable rollouts and preserves the scene-level separation outcome relative to manual initialization; this test evaluates initialization consistency rather than material-identification accuracy. In our setting, the material prior affects the pre-latch dynamics and failure time, but the post-latch transfer behavior is governed by identity-conditioned routing and is therefore not specific to how parameters are initialized.

\section{Limitations}

\noindent \textbf{(L1) Binary split only.} Our current formulation targets a persistent binary split and does not model crack growth, branching, or dynamic fragment creation. Extending the approach to multiple fragments would require sparse per-node field allocation rather than dense two-field storage.

\noindent \textbf{(L2) Simplified contact model.} Contact handling is intentionally lightweight: we use a fixed latched normal and a frictionless normal impulse. This approximation is most appropriate for approximately planar interfaces and may degrade under large rotations or highly curved fracture surfaces.

\noindent \textbf{(L3) Fragment--scene contact robustness.} We focus on fragment-conditioned transfer and inter-fragment re-contact. In the Ficus static-plane experiment (Table~\ref{tab:scene_interaction}, Panel~A), Full is not classified as successful under the strict identity-crossing criterion, indicating that robust fragment--scene contact remains an open problem.

\noindent \textbf{(L4) Memory overhead and appearance.} The dense two-field implementation doubles the branch-specific grid-buffer footprint. Newly exposed fracture surfaces also lack explicit geometry and texture because the input Gaussian reconstruction does not observe object interiors.

\noindent These limitations bound our claims to post-event transfer topology and controlled fragment--fragment contact consistency.

\section{Conclusion}

We identified post-fracture coupling in Gaussian--MPM as a mismatch between disconnected structural topology and a single-valued grid velocity field, which induces cross-fragment momentum leakage through shared normalization. \method resolves this mismatch by constructing fragment-conditioned nodal states in a single P2G traversal, preserving identity through field-aware G2P, and applying momentum-conserving contact only for approaching fragment fields. Experiments show that fragment-conditioned routing eliminates realized cross-fragment mixing by construction and improves post-fracture separation behavior, while the contact projection reduces penetration during controlled re-contact. Overall, our results support a simple design principle for physics-ready 3D representations: when topology disconnects, the simulator should change the local dynamical state—not only the constitutive stress.

Future work includes sparse multi-fragment field allocation, enhanced frictional and fragment–scene contact, and tighter integration with crack growth and interior/appearance completion for realistic broken-surface rendering.

\bibliography{aaai2027}

\end{document}